\documentclass{article}

\usepackage[accepted]{icml2023}
\usepackage{microtype}
\usepackage{graphicx}
\usepackage{subfigure}
\usepackage{booktabs} 

\usepackage{lscape}
\usepackage{array}
\usepackage{longtable}
\usepackage{multirow}
\usepackage{times}
\usepackage{hyperref}

\usepackage{amsmath}
\usepackage{amssymb}
\usepackage{mathtools}
\usepackage{amsthm}
\usepackage[cjk]{kotex}
\usepackage[normalem]{ulem}
\useunder{\uline}{\ul}{}

\usepackage[capitalize,noabbrev]{cleveref}

\theoremstyle{plain}

\theoremstyle{definition}

\theoremstyle{remark}

\usepackage[textsize=tiny]{todonotes}

\icmltitlerunning{Submission and Formatting Instructions for ICML 2023}

\begin{document}

\twocolumn[
\icmltitle{Knowledge Graph-Augmented Korean Generative Commonsense Reasoning}



\icmlsetsymbol{equal}{*}

\begin{icmlauthorlist}
\icmlauthor{Dahyun Jung}{yyy}
\icmlauthor{Jaehyung Seo}{yyy}
\icmlauthor{Jaewook Lee}{yyy}
\icmlauthor{Chanjun Park}{yyy,yyyc}
\icmlauthor{Heuiseok Lim}{yyy}
\end{icmlauthorlist}

\icmlaffiliation{yyy}{Department of Computer Science and Engineering, Korea University, Seoul 02841, Korea}
\icmlaffiliation{yyyc}{Upstage, Gyeonggi-do, Korea}


\icmlcorrespondingauthor{Heuiseok Lim}{limhseok@korea.ac.kr}

\icmlkeywords{Machine Learning, ICML}

\vskip 0.3in
]



\printAffiliationsAndNotice{}  

\begin{abstract}
Generative commonsense reasoning refers to the task of generating acceptable and logical assumptions about everyday situations based on commonsense understanding. By utilizing an existing dataset such as Korean CommonGen, language generation models can learn commonsense reasoning specific to the Korean language. However, language models often fail to consider the relationships between concepts and the deep knowledge inherent to concepts. To address these limitations, we propose a method to utilize the Korean knowledge graph data for text generation. Our experimental result shows that the proposed method can enhance the efficiency of Korean commonsense inference, thereby underlining the significance of employing supplementary data.
\end{abstract} 

\section{Introduction}
Generative commonsense reasoning is a constrained text generation task that enables language models to learn the capacity to generate text while considering commonsense information~\cite{lin2019commongen}. 
This task requires making a sentence describing a commonplace scene using a set of given concepts. For instance, given a set of concepts like {toothbrush, cup, placed, toothpaste}, the models should be able to output a sentence like ``A toothbrush and toothpaste are placed in a cup.'' 
While humans generally possess this commonsense knowledge, it's not intrinsically present within language models, which emphasizes the importance of this task~\cite{davis2015commonsense}.

\begin{figure}[t]
\centering
\includegraphics[width=\columnwidth]{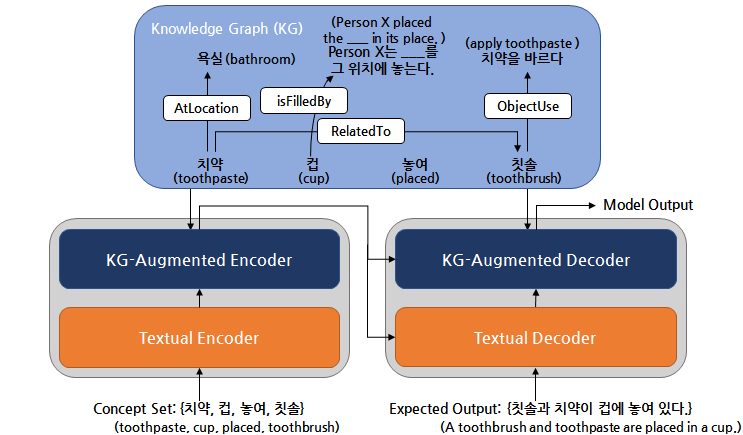}
\caption{The illustration of knowledge graph-augmented model}
\label{fig:figure}
\end{figure}

Korean CommonGen~\cite{seo-etal-2022-dog} is a dataset that constructs on CommonGen~\cite{lin2019commongen}, the generative commonsense reasoning dataset, in Korean. Tasks that generate text using this dataset can consider commonsense information specific to the Korean language. 
This dataset can be harnessed to learn commonsense knowledge that incorporates the linguistic and cultural features of the Korean language.
However, the limitation of the tasks is its inability to capture the relationships between concepts absent from the dataset. 
The dataset does not cover every possible concept or scenario, leading to gaps in the commonsense knowledge that the model can learn. 
Additionally, the dataset lacks depth in certain areas, potentially leading to a superficial understanding of concepts.

In this study, we mitigate the limitation by proposing a method to augment commonsense information in generative commonsense reasoning tasks. By leveraging the Korean commonsense knowledge graph, Ko-ATOMIC, and applying it to text generation, we can take into account not only the existing commonsense relationships between concepts, but also the deeper knowledge embedded in them. We construct a model that can maximize the utilization of this knowledge graph and conduct experiments. The results intuitively demonstrate that our approach is suitable for the task of generative commonsense reasoning in Korean. 
This research lays the foundation for leveraging language-specific resources to enhance the capability of artificial intelligence.

\begin{table*}[]
\renewcommand{\arraystretch}{1.2}
\centering
{\small 
\begin{tabular}{l|ccccccc}
\hline
\textbf{Model} & \textbf{BLEU-3} & \textbf{BLEU-4} & \textbf{ROUGE-2} & \textbf{ROUGE-L} & \textbf{METEOR} & \textbf{mBERTScore} & \textbf{KoBERTScore} \\ \hline
KoBART   & 39.54          & 29.16          & 53.6           & 68.55       & 51.17      & \textbf{87.41} & {\ul 92.59} \\
mBART    & \textbf{41.83} & \textbf{31.63} & \textbf{54.21} & 68.36       & 52.08      & 87.25          & 92.26       \\
mBART-50 & 40.51          & 30.2           & 53.5           & 68.18       & 50.9       & 87.31          & 92.26       \\ \hline
KoBART (Ours)         & {\ul 41.13}     & {\ul 30.23}     & {\ul 54.02}      & \textbf{70.1}    & \textbf{57.62}  & {\ul 87.33}         & \textbf{92.79}       \\
mBART (Ours)  & 36.16          & 25.55          & 50.69          & 67.83       & 54.78      & 86.36          & 90.58       \\
mBART-50 (Ours) & 36.85          & 26.26          & 51.91          & {\ul 68.67} & {\ul 56.2} & 86.77          & 90.83       \\ \hline
\end{tabular}
}
\caption{Automatic evaluation of generation quality. The first group
of models is baseline models, while the second group is our proposed knowledge graph-augmented models. The best models are \textbf{bold}, and the second best ones are {\ul underlined} within each metric.}
\label{tab:table}
\end{table*}
\section{Knowledge Graph-Augmented Commonsense Reasoning}

\subsection{Data Construction}
\noindent\textbf{Text generation data for commonsense reasoning}
We use the Korean CommonGen data for generative commonsense inference in Korean. 
This data considers the linguistic characteristics of the Korean language, including uniquely Korean socio-cultural terms.

\noindent\textbf{Commonsense knowledge graph}
The most well-known existing commonsense-based knowledge graph is ATOMIC~\cite{sap2019atomic}. In this study, we utilize the Ko-ATOMIC~\footnote{https://github.com/jooinjang/Ko-ATOMIC} graph, which is a translation of the existing English-based ATOMIC.

\subsection{Enhancing Model with Knowledge}
We aim to improve commonsense reasoning by utilizing the methodology of KG-BART~\cite{liu2021kg}, a knowledge graph augmented pre-trained language generation model. 

The KG-BART model is differentiated from previous pre-trained language generation models by incorporating a knowledge graph, an important source of rich relational information between commonsense concepts. By utilizing this knowledge graph, the model is able to incorporate complex relationships between concepts into the text generation process. 
This implies that the generated sentences are not only grounded in learned linguistic patterns, but are also infused with structured knowledge, resulting in more logical text. 

KG-BART consists of an encoder-decoder architecture that takes text concepts and knowledge graph as input (Figure~\ref{fig:figure}).
The encoder and decoder are supplemented with a KG-augmented Transformer~\cite{vaswani2017attention} module based on a graph attention mechanism for incorporating entity-oriented knowledge information into the token representation.
This feature allows models to capture the inherent structural correlations of intra-concept and inter-concept in the graph. 

\subsection{Results}
We experiment with an improved methodology by applying a Korean knowledge graph.  
The metrics include BLEU~\cite{papineni-etal-2002-bleu}, ROUGE~\cite{lin-2004-rouge}, METEOR~\cite{banerjee-lavie-2005-meteor}, and BERTScore~\cite{zhang2019bertscore} using KoBERT~\footnote{https://github.com/SKTBrain/KoBERT} and mBERT~\cite{libovicky2019language}. We compare the baseline models, KoBART~\footnote{https://github.com/SKT-AI/KoBART}, mBART~\cite{liu2020multilingual}, with the model enhanced by our proposed approach.
mBART is a sequence-to-sequence encoder pre-trained on large corpora in multiple languages using BART~\cite{lewis2019bart}. 
KoBART is trained on a large corpus of millions of Korean sentences collected from Wikipedia, news, etc. 
mBART-50 is pre-trained for 50 languages using mBART's checkpoints. 

In Table \ref{tab:table}, we present the results of the automatic evaluation of generation quality for different models.
The model that incorporates the knowledge graph augmentation shows a noticeable improvement in several metrics. Our improved KoBART model yields significantly better results in the ROUGE-L. It also surpasses all other models in the METEOR and KoBERTScore, achieving 57.62 and 92.79, respectively. This suggests that the model's language generation quality is highly accurate. Our proposed mBART and mBART-50 models also demonstrate a substantial increase in their performance. Even though they didn't achieve the highest scores, the improvements in their metrics are promising, highlighting the potential and effectiveness of the structured data-enhanced approach.
In conclusion, our knowledge graph-augmented models exhibit notable improvements over the baseline models in several evaluation metrics, underlining the viability of our proposed method in enhancing the quality of the generated text. 

\section{Conclusion}
In this paper, we explored the utility of the Korean commonsense knowledge graph in advancing the ability of commonsense reasoning in the Korean language. By considering relationships and in-depth knowledge, we managed to illustrate an improvement in inference quality. We demonstrated that using these graphs as an adjunct to commonsense reasoning systems can help bridge the gaps in knowledge. 

\section*{Acknowledgements}
This research was supported by the Core Research Institute Basic Science Research Program through the National Research Foundation of Korea(NRF) funded by the Ministry of Education(NRF-2021R1A6A1A03045425). This work was supported by Institute for Information \& communications Technology Planning \& Evaluation(IITP) grant funded by the Korea government(MSIT) (No. 2022-0-00369, (Part 4) Development of AI Technology to support Expert Decision-making that can Explain the Reasons/Grounds for Judgment Results based on Expert Knowledge).

\nocite{langley00}

\bibliography{example_paper}
\bibliographystyle{icml2023}

\end{document}